\documentclass[10pt,twocolumn,letterpaper]{article}

\usepackage{cvpr}
\usepackage{times}
\usepackage{epsfig}
\usepackage{graphicx}
\usepackage{amsmath}
\usepackage{amssymb}
\usepackage{subfigure}
\usepackage{overpic}

\usepackage{enumitem}
\setenumerate[1]{itemsep=0pt,partopsep=0pt,parsep=\parskip,topsep=5pt}
\setitemize[1]{itemsep=0pt,partopsep=0pt,parsep=\parskip,topsep=5pt}
\setdescription{itemsep=0pt,partopsep=0pt,parsep=\parskip,topsep=5pt}

\usepackage[pagebackref=true,breaklinks=true,letterpaper=true,colorlinks,bookmarks=false]{hyperref}

\cvprfinalcopy 

\def\cvprPaperID{} 

\graphicspath{{figures/}}

\begin{document}

\newcommand{\figref}[1]{图\ref{#1}}
\newcommand{\tabref}[1]{表\ref{#1}}
\newcommand{\equref}[1]{式\ref{#1}}
\newcommand{\secref}[1]{第\ref{#1}节}


\title{Target-less registration of point clouds: A review}

\author{Yue Pan\\
ETH Zurich, D-BAUG \\
{yuepan@student.ethz.ch}
}
\maketitle

\begin{abstract}
Point cloud registration has been one of the basic steps of point cloud processing, which has a lot of applications in remote sensing and robotics. In this report, we summarized the basic workflow of target-less point cloud registration,namely correspondence determination and transformation estimation. Then we reviewed three commonly used groups of registration approaches, namely the feature matching based methods, the iterative closest points algorithm and the randomly hypothesis and verify based methods. Besides, we analyzed the advantage and disadvantage of these methods are introduced their common application scenarios. At last, we discussed the challenges of current point cloud registration methods and  proposed several open questions for the future development of automatic registration approaches.

\end{abstract}

\section{Introduction}

In recent decades, point cloud has become a more and more common representation of the 3D world. Point cloud collected by laser scanner or RGBD-cameras can be used for landslide monitoring \cite{rowlands2003landslide}, solar potential analysis \cite{solar}, three dimensional model reconstruction \cite{citymodel}, cultural heritage protection \cite{Heritage}, forest management \cite{Forestreg}\cite{forestreg2}, robot ego-localization \cite{loam} and high definition maps production for self-driving cars \cite{hdmapping}. 

\begin{figure}[!h]
\begin{center}
\includegraphics[width=0.8\linewidth]{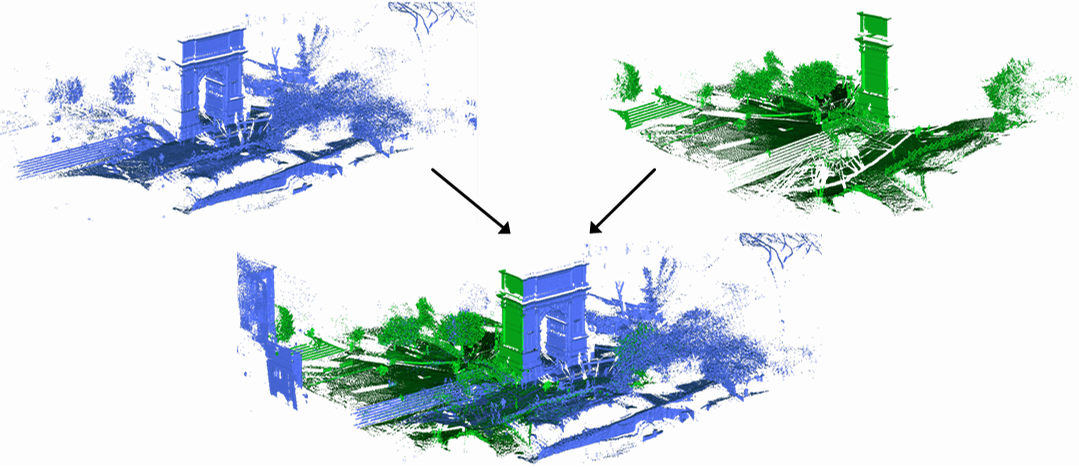}
\end{center}
   \caption{An example of registration of laser scans \cite{globallyreg}}
\label{img1}
\end{figure}

The issue is that the collection of point cloud is limited to the perspective so that it is not possible to achieve the 360 degree complete sampling of the target object's surface from a single viewpoint (station). Usually, we would set up several stations around the target object to have a throughout scan. However, these scans from different viewpoints are in their corresponding station center coordinate system. To unify these scans into a common mapping coordinate system, we should accomplish the so-called point cloud registration procedure, as shown in Fig.\ref{img1}.

As the basic step of point cloud processing and the prerequisites of segmentation, classification and 3D model reconstruction, point cloud registration plays an important role in various of remote sensing and robotics applications. 
By adopting point cloud registration to get the transformation between two adjacent frames (scans), we can get the change of pose of a robot or an unmanned vehicle. This is called the LiDAR odometry , which is a heated topic in Simultaneous Localization and Mapping (SLAM) technology.

The traditional solution to point cloud registration is using some highly-reflective targets as the tie points for coordinate system transformation. Since this solution still needs the assist of artificial targets and the manual pining of targets in the point clouds, the process is labor-consuming and time-consuming. To automate the process, over the past twenty years, plenty of target-less point cloud registration approaches have been proposed in the fields of remote sensing and computer vision so as to automatically register point clouds together.

In literature, the task of point cloud registration generally follows a two-step workflow: determine correspondences and then estimate the transformation. The first step is correspondences determination. The correspondences can be geometric primitives like points, lines, planes and even specific objects. As the preparation, we usually need to detect the key points, fit the key lines, planes or extract the specific objects. Then we can extract the neighborhood feature and match those geometric primitives according to the feature similarity. Alternatively, the geometric or adjacency relationship can be adopted to get correspondences. Besides, we can keep randomly sampling a minimum set of correspondences and finally choose the set which leads to the transformation with the largest number of inliers.

The second step is transformation estimation. Given the correspondences, our goal is to solve the transformation (namely, translation and rotation) between two point clouds.  Generally, we firstly define a reasonable target function with regard to the transformation parameters. It guarantees a good registration result when the function’s value is minimized. Then we can minimize (optimize) the target function using methods like Singular Value Decomposition (SVD), Linear Least Square (LLS) and also non-linear optimization algorithms such as Gauss-Newton and Levenberg-Marquardt. The transformation parameters corresponding to the minimal target function value are what we’d like to achieve.

The closely related studies would be briefly reviewed and discussed as follows. In section \ref{sec2}, common transformation estimation methods shared by various registration algorithm would be reviewed. In section \ref{sec3}, \ref{sec4} and \ref{sec5}, registration methods based on feature matching, iterative closest points and randomized hypothesize-and-verify would be reviewed respectively. Section \ref{sec6} consists of the summary of introduced algorithms and the outlook of existing challenges and open questions.

\section{Transformation estimation}\label{sec2}

\subsection{Target function}
 
\begin{figure*}[!t]
\begin{center}
\includegraphics[width=0.8\linewidth]{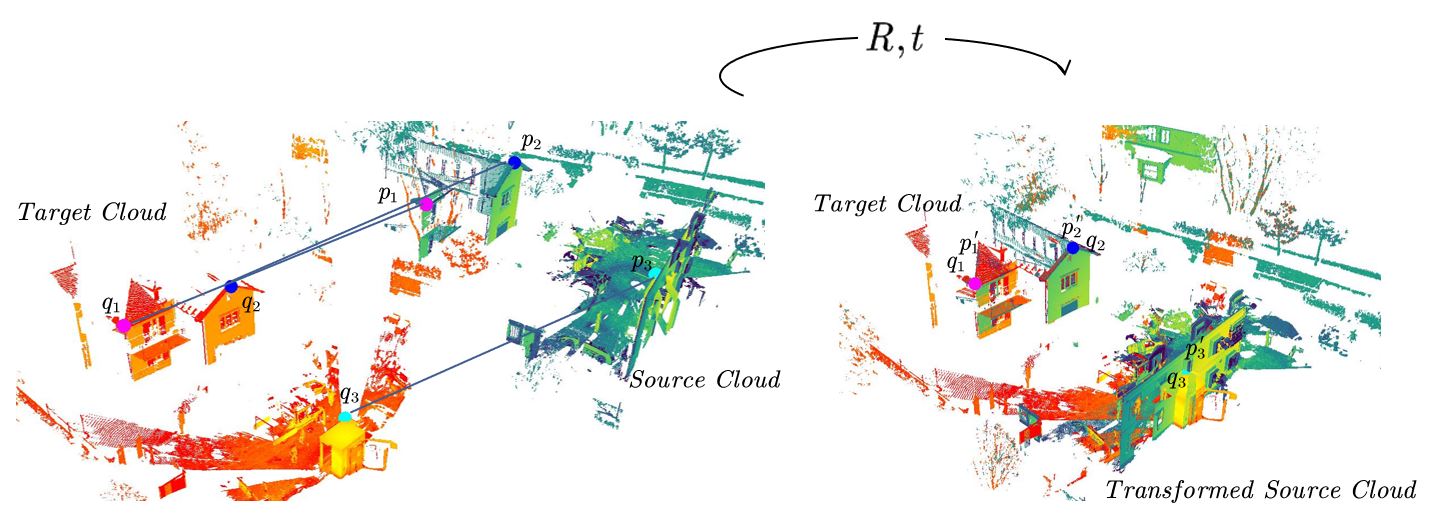}
\end{center}
   \caption{An example of transformation estimation with point correspondences.}
\label{img2}
\end{figure*}

 Given the correspondence points $p_i,q_i$ in the source (moving) point cloud and the target (referenced) point cloud, we'd like to estimate the transformation $\left\{ R^*,t^* \right\}$ from the source to the target point cloud, as shown in Fig.\ref{img2}. the target function under point-to-point distance metric can be drawn as Eq.\ref{equ1-1}, which leads to minimum sum of distance between correspondences after registration. In this case, at least three pair of correspondence are needed.

\begin{equation}
  \left\{ R^*,t^* \right\} =\underset{\left\{ R,t \right\}}{\text{arg}\min}\left( \sum_i{\lVert Rp_i+t-q_i \rVert ^2} \right) 
    \label{equ1-1}
\end{equation}

\begin{figure*}[!h]
\begin{center}
\includegraphics[width=0.8\linewidth]{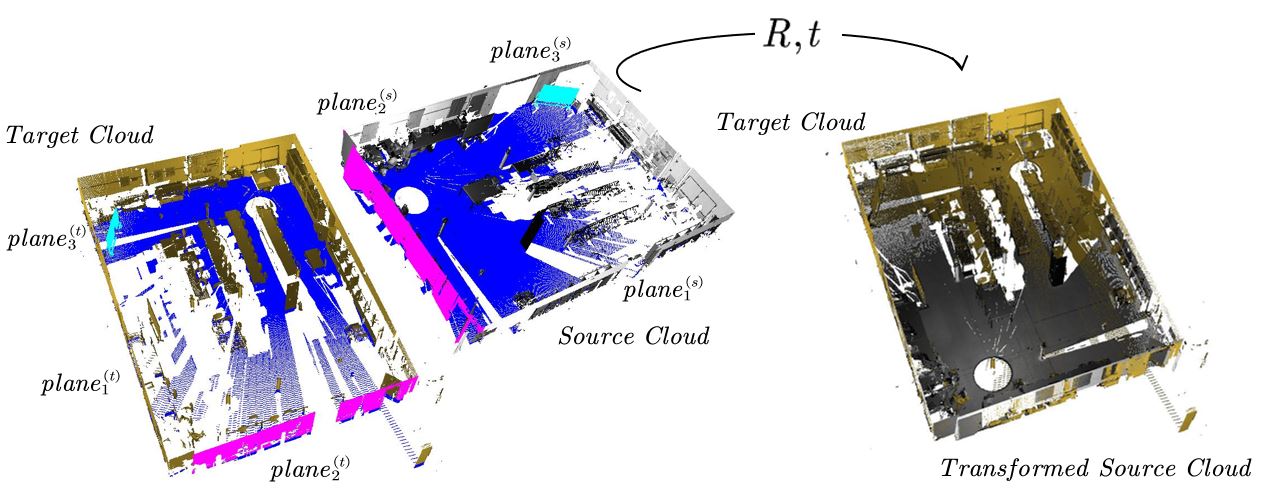}
\end{center}
   \caption{An example of transformation estimation with plane correspondences.}
\label{img3}
\end{figure*}

If the geometric primitive used is planes instead of points as shown in Fig.\ref{img3}, the target function goes like Eq.\ref{equ1-2}. When fitting the planes, we can get the normal vector $n$ and the distance from coordinate origin to the plane $\rho$. Registration's target is to minimize the sum of difference of the normal vector and the distance between corresponding planes after updating the estimated transformation. In this case, still at least three pair of corresponding planes are needed.

\begin{equation}
\left\{ R^*,t^* \right\} =\underset{\left\{ R,t \right\}}{\text{arg}\min}\left( \sum_i{\lVert \left[ \begin{array}{c}
	Rn_{i}^{\left( s \right)}-n_{i}^{\left( t \right)}\\
	\rho _{i}^{\left( s \right)}-\rho _{i}^{\left( t \right)}+\left( Rn_{i}^{\left( s \right)} \right) ^Tt\\
\end{array} \right] \rVert ^2} \right) 
    \label{equ1-2}
\end{equation}

\subsection{Singular Value Decomposition (SVD)}

A popular closed-form solution to point-to-point target function Eq.\ref{equ1-1} is the method based on SVD\cite{svd}. Firstly, we calculate the centroids of the source and target point clouds (Eq.\ref{equ2}) and the decentralized coordinate of all the correspondences (Eq.\ref{equ3}). After that, we apply SVD (Eq.\ref{equ4}) and get the rotation matrix $R$ and translation vector $t$ from the decomposed matrices as Eq.\ref{equ5}.

\begin{equation}
\bar{p}=\frac{1}{N}\sum_{i=1}^N{p_i},\bar{q}=\frac{1}{N}\sum_{i=1}^N{q_i}
    \label{equ2}
\end{equation}

\begin{equation}
p_{i}^{'}=p_i-\bar{p},\ q_{i}^{'}=q_i-\bar{q}
    \label{equ3}
\end{equation}

\begin{equation}
U\varSigma V^T=\sum_{i=1}^N{p_{i}'q_{i}'^T}
    \label{equ4}
\end{equation}

\begin{equation}
R^*=VU^T,\ t^*=\bar{q}-R^*\bar{p}
    \label{equ5}
\end{equation}

\subsection{Linear Least Square (LLS)}
Another popular solution to Eq.\ref{equ1-1} is linear least square parameter estimation \cite{lls}. Since we can do the approximation $\sin \left( \alpha \right) \approx \alpha \ when\ \alpha \rightarrow 0$, the rotation matrix can be represented as Eq.\ref{equ6}. Then we can construct the observation function as Eq.\ref{equ7}, which can be arranged as Eq.\ref{equ8}. Since the design matrix $A$ and observation vector $l$ can be calculated, the transformation parameters $\hat{x}$ can then be estimated as Eq.\ref{equ9}. The rotation matrix is then restored from the rotation vector.

\begin{equation}
R\approx \left[ \begin{matrix}
	1&		-\gamma&		\beta\\
	\gamma&		1&		-\alpha\\
	-\beta&		\alpha&		1\\
\end{matrix} \right] =\left[ \begin{array}{c}
	\alpha\\
	\beta\\
	\gamma\\
\end{array} \right] _{\times}+I
    \label{equ6}
\end{equation}

\begin{equation}
v_i=\left[ \begin{matrix}
	1&		-\gamma&		\beta\\
	\gamma&		1&		-\alpha\\
	-\beta&		\alpha&		1\\
\end{matrix} \right] \left[ \begin{array}{c}
	x_{i}^{\left( p \right)}\\
	y_{i}^{\left( p \right)}\\
	z_{i}^{\left( p \right)}\\
\end{array} \right] +\left[ \begin{array}{c}
	t_x\\
	t_y\\
	t_z\\
\end{array} \right] -\left[ \begin{array}{c}
	x_{i}^{\left( q \right)}\\
	y_{i}^{\left( q \right)}\\
	z_{i}^{\left( q \right)}\\
\end{array} \right] 
    \label{equ7}
\end{equation}

\begin{equation}
\begin{split}
\underset{v_i}{\underbrace{\left[ \begin{array}{c}
	v_{i}^{\left( x \right)}\\
	v_{i}^{\left( y \right)}\\
	v_{i}^{\left( z \right)}\\
\end{array} \right] }}=&\underset{A_i}{\underbrace{\left[ \begin{matrix}
	0&		z_{i}^{\left( p \right)}&		-y_{i}^{\left( p \right)}&		1&		0&		0\\
	-z_{i}^{\left( p \right)}&		0&		x_{i}^{\left( p \right)}&		0&		1&		0\\
	y_{i}^{\left( p \right)}&		-x_{i}^{\left( p \right)}&		0&		0&		0&		1\\
\end{matrix} \right] }}\underset{x}{\underbrace{\left[ \begin{array}{c}
	\alpha\\
	\beta\\
	\gamma\\
	t_x\\
	t_y\\
	t_z\\
\end{array} \right] }}
\\
-&\underset{l}{\underbrace{\left[ \begin{array}{c}
	x_{i}^{\left( q \right)}-x_{i}^{\left( p \right)}\\
	y_{i}^{\left( q \right)}-y_{i}^{\left( p \right)}\\
	z_{i}^{\left( q \right)}-z_{i}^{\left( p \right)}\\
\end{array} \right] }}
\end{split}
    \label{equ8}
\end{equation}
\begin{equation}
\hat{x}=\left[ \begin{matrix}
	\alpha&		\beta&		\gamma&		t_x&		t_y&		t_z\\
\end{matrix} \right] ^T=\left( A^TA \right) ^{-1}A^Tl
    \label{equ9}
\end{equation}

\subsection{Three plane correspondence method}

As for the plane-to-plane target function Eq.\ref{equ1-2}, a simple solution is the three plane plus one intersection point method \cite{visualplanar}. The selected planes have to be linearly independent and intersect at a unique point in order for the transformation parameters to be fully recovered. We can calculate rotation matrix from normal vectors as Eq.\ref{equ10} and Eq.\ref{equ11}. The intersection point is calculated as Eq.\ref{equ12}. The translation vector is then calculated from the vector between the corresponding intersection point of these planes as Eq.\ref{equ13}.  

\begin{equation}
getRotation\left( v_1,v_2 \right) =I+\left[ v_1\times v_2 \right] _{\times}+\left[ v_1\times v_2 \right] _{\times}^{2}\frac{1-v_1\cdot v_2}{\lVert v_1\times v_2 \rVert}
    \label{equ10}
\end{equation}

\begin{equation}
\begin{split}
    R_1=&getRotation\left( n_{1}^{\left( s \right)},n_{1}^{\left( t \right)} \right) 
\\
R_2=&getRotation\left( R_1n_{2}^{\left( s \right)},n_{2}^{\left( t \right)} \right) 
\\
R_3=&getRotation\left( R_2R_1n_{3}^{\left( s \right)},n_{3}^{\left( t \right)} \right) 
\\
R^*=&R_3R_2R_1
\end{split}
    \label{equ11}
\end{equation}

\begin{equation}
x_{int}^{\left( t \right)}=\left[ \begin{matrix}
	a_{1}^{\left( t \right)}&		b_{1}^{\left( t \right)}&		c_{1}^{\left( t \right)}\\
	a_{2}^{\left( t \right)}&		b_{2}^{\left( t \right)}&		c_{2}^{\left( t \right)}\\
	a_{3}^{\left( t \right)}&		b_{3}^{\left( t \right)}&		c_{3}^{\left( t \right)}\\
\end{matrix} \right] ^{-1}\left[ \begin{array}{c}
	-d_{1}^{\left( t \right)}\\
	-d_{2}^{\left( t \right)}\\
	-d_{3}^{\left( t \right)}\\
\end{array} \right] 
    \label{equ12}
\end{equation}

\begin{equation}
t^*=x_{int}^{\left( t \right)}-x_{int}^{\left( s \right)}
    \label{equ13}
\end{equation}

\section{Feature matching based algorithms}\label{sec3}

\subsection{Feature matching workflow}

\begin{figure}[!ht]
\begin{center}
\includegraphics[width=1.0\linewidth]{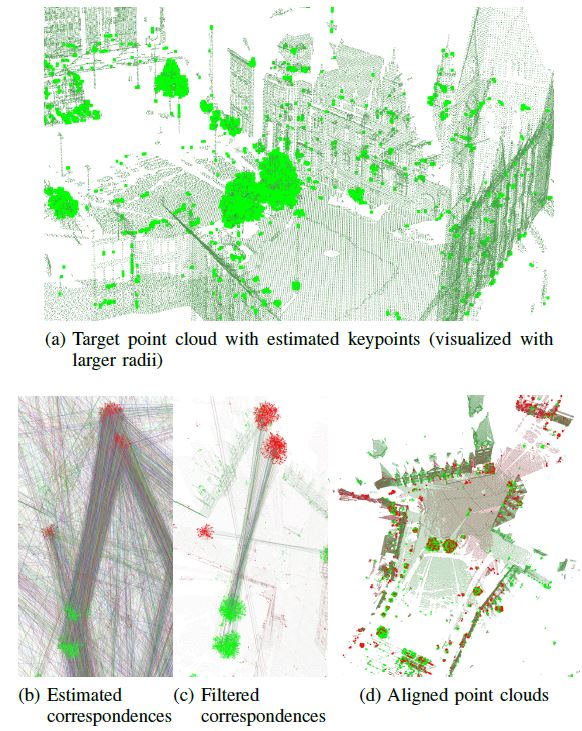}
\end{center}
   \caption{An example of feature matching based registration \cite{pclreg}.}
\label{img4}
\end{figure}

Most of the feature matching based registration algorithms (as shown in Fig.\ref{img4}) follow the similar workflow. 

Firstly, keypoint detectors such as intrinsic shape signature \cite{intrinsic}, 3D Harris \cite{3dharris} and local curvature maximum are explored to detect keypoints from original point clouds. These keypoints are more significant geometrically so that more representative features can be extracted from them. 

\begin{figure}[!h]
\begin{center}
\includegraphics[width=1.0\linewidth]{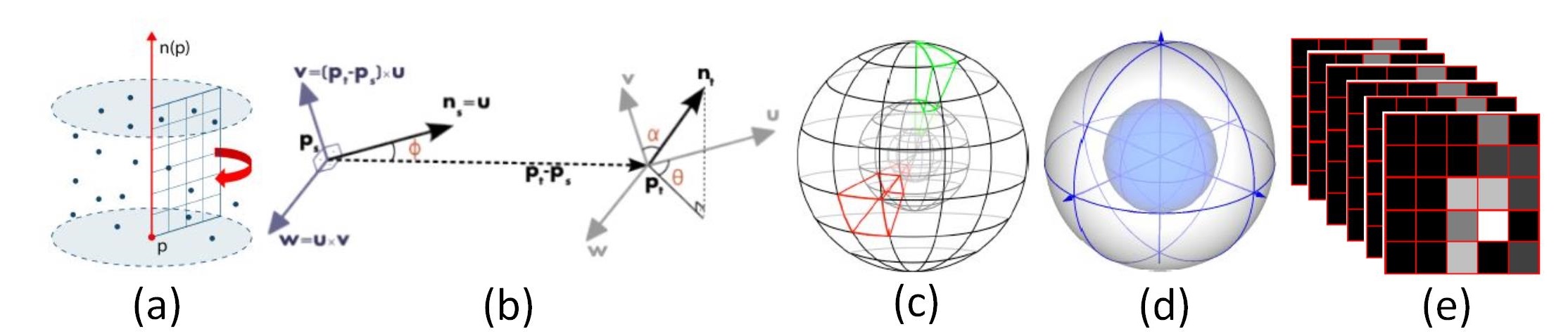}
\end{center}
   \caption{Popular handcrafted point features: (a)Spin Images (SI)\cite{SI}, (b)Fast Point Feature Histogram (FPFH)\cite{fpfhdes}, (c)3D Shape Context (3DSC)\cite{3dsc}, (d)SHOT descriptor\cite{Shot} (e)Binary Shape Context (BSC) \cite{BSC}}
\label{img5}
\end{figure}

Secondly, local feature descriptors such as Spin Image \cite{SI}, Fast Point Feature Histograms (FPFH) \cite{fpfhdes}, SHOT descriptor \cite{Shot}, Rotational Projection Statistics (RoPS)\cite{rops} , 3D Shape Context \cite{3dsc} and Binary Shape Context \cite{BSC} are generated to encode the local neighborhood information of each keypoint. These feature descriptors should be invariant or insensitive to rigid transformation (translation and rotation) and have high precision and recall for matching. Several popular handcrafted feature descriptors are shown in Fig.\ref{img5}.

Recently, apart from these manually craft features, there are some learned features using deep neural network. A state-of-art point-based model is the PointNet \cite{Pointnet}, which is able to learn the descriptive point-wise feature of the point cloud for classification and semantic segmentation. A better network structure for point feature extraction used for matching and registration is the so-called siamese network with triplet loss function. An example is the Perfect Match Net \cite{perfectmatch}, which outperforms all the existing handcrafted and learned features on matching accuracy and efficiency with only about 16 dimensional learned feature. Since deep learning has already proved its superiority to traditional methods on both 2D and 3D computer vision, point cloud registration based on deep learning may finally be the main stream solution in the near future.

\begin{figure}[!h]
\begin{center}
\includegraphics[width=1.0\linewidth]{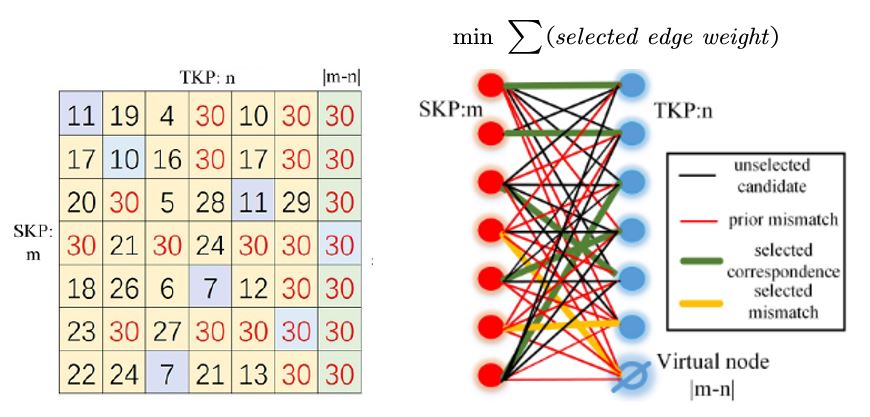}
\end{center}
   \caption{Feature matching by Bipartite Graph minimum weight match\cite{IGSP}.}
\label{img6}
\end{figure}

Thirdly, various feature matching strategies such as reciprocal nearest neighbor, nearest neighbor similarity ratio test and bipartite graph minimal match \cite{IGSP} as shown in Fig.\ref{img6} are adopted to identify the initial match. 

\begin{figure}[!h]
\begin{center}
\includegraphics[width=1.0\linewidth]{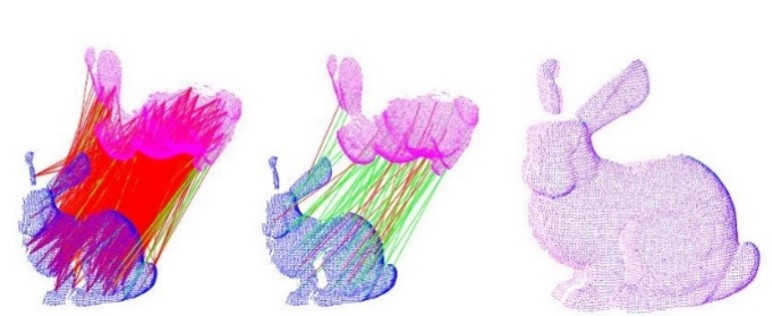}
\end{center}
   \caption{An example of feature matching and correspondence filter \cite{GORE}}
\label{img7}
\end{figure}

However, there may still be a lot of outlier matches (red lines in Fig.\ref{img7}) among them . Then the incorrect correspondences are eliminated based on methods such as RANSAC \cite{RANSAC}, geometric consistency constraints \cite{yangreg} or Game Theory based matching algorithm \cite{gametheory}. The RANSAC based on correspondence selects the best correspondence triplet which gives rise to most inliers after transformation. 

Finally, the spatial transformation between the point cloud pair is estimated based on the correspondences after filtering.

\subsection{Analysis and evaluation}
Feature matching based registration methods are global registration approaches because there's no requirement of transformation initial guess. The drawback is that they are not accurate enough since they are based on keypoints instead of the denser raw point cloud so that they are often regarded as coarse registration \cite{coarsereg}. Only with proper matching strategy and outlier filter, can feature matching based methods be robust to noise, occlusion and low overlapping ratio. Besides, these methods are usually time-consuming due to their complex feature extraction, matching and filtering procedure.

\section{Iterative Closest Points based algorithms}\label{sec4}

\subsection{Classic ICP}

\begin{figure}[!h]
\begin{center}
\includegraphics[width=1.0\linewidth]{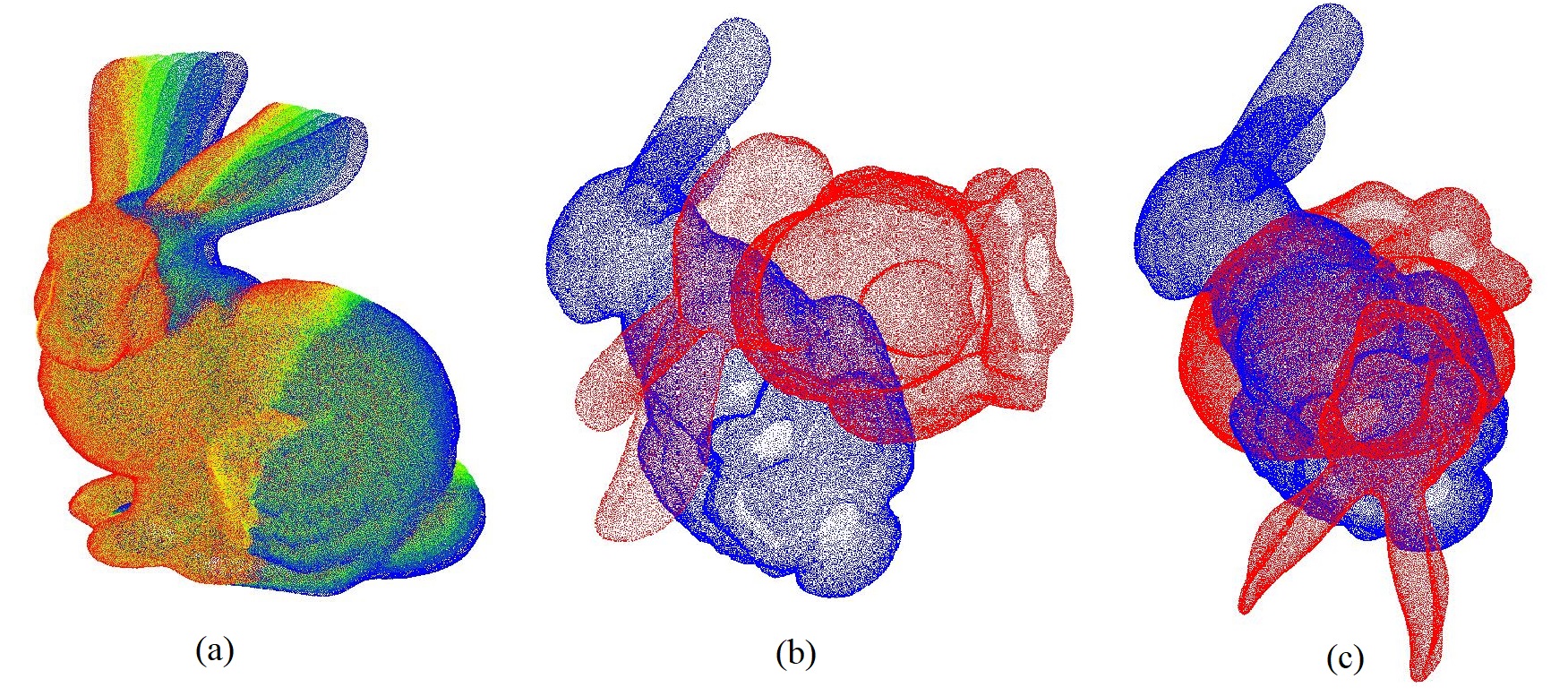}
\end{center}
   \caption{Example of ICP algorithm: (a) Successful registration of different iterations, (b) bad initial guess with too large rotation difference, (c) failed registration result of ICP with bad initial guess.  }
\label{img8}
\end{figure}

The Iterative Closest Point (ICP) algorithm \cite{ICP} is the most commonly used fine registration method due to its conceptual simplicity and high usability. With a good initial transformation, ICP accomplishes a locally optimal registration by alternately solving for point-to-point closest correspondences and optimal rigid transformation until convergence. At the correspondence determination step, ICP simply assumes closest points of source and target point cloud as correspondences. Then the transformation is estimated from the closest points by minimizing Eq.\ref{equ1-1}. The source point cloud is updated with the transformation and the new closest point correspondences can be calculated so that the aforementioned process can be done iteratively, as shown in Fig.\ref{img8}(a).

\subsection{ICP variants}

The variants of ICP mainly focus on different processing steps (correspondence determination, outlier correspondence rejection and transformation estimation target function construction) of classic ICP algorithm \cite{ICPcompare}.

\begin{figure}[!ht]
\begin{center}
\includegraphics[width=1.0\linewidth]{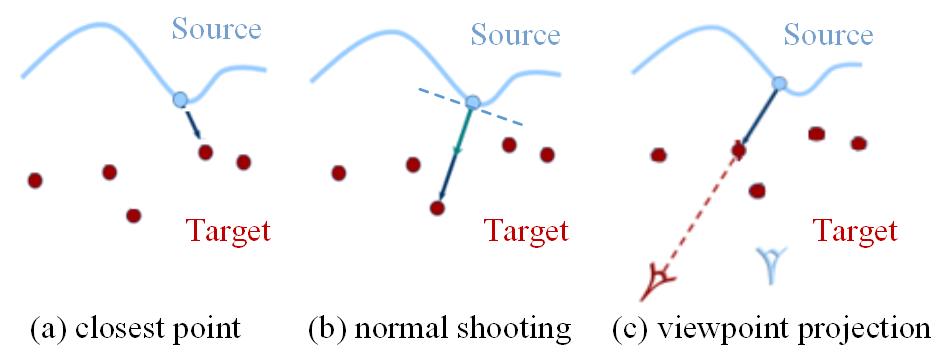}
\end{center}
   \caption{ICP variants with different correspondence determining principle.}
\label{img9}
\end{figure}

For correspondence determination, as shown in Fig.\ref{img9}, there are alternative principle like normal shooting which is suitable for registration of smooth structure and viewpoint projection which is more efficient when the viewpoint is already known.

\begin{figure}[!ht]
\begin{center}
\includegraphics[width=1.0\linewidth]{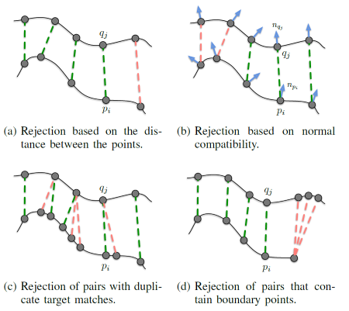}
\end{center}
   \caption{ICP variants with different correspondence rejection methods \cite{pclreg}.}
\label{img10}
\end{figure}

As for the outlier correspondence rejection, as shown in Fig.\ref{img10}, we can set the correspondence distance threshold , normal vector compatibility and matching uniqueness to get rid of correspondence outliers. \cite{trimmed}
 propose the Trimmed-ICP algorithm which estimates the distance threshold according to the approximate overlapping ratio.

\begin{figure}[!h]
\begin{center}
\includegraphics[width=1.0\linewidth]{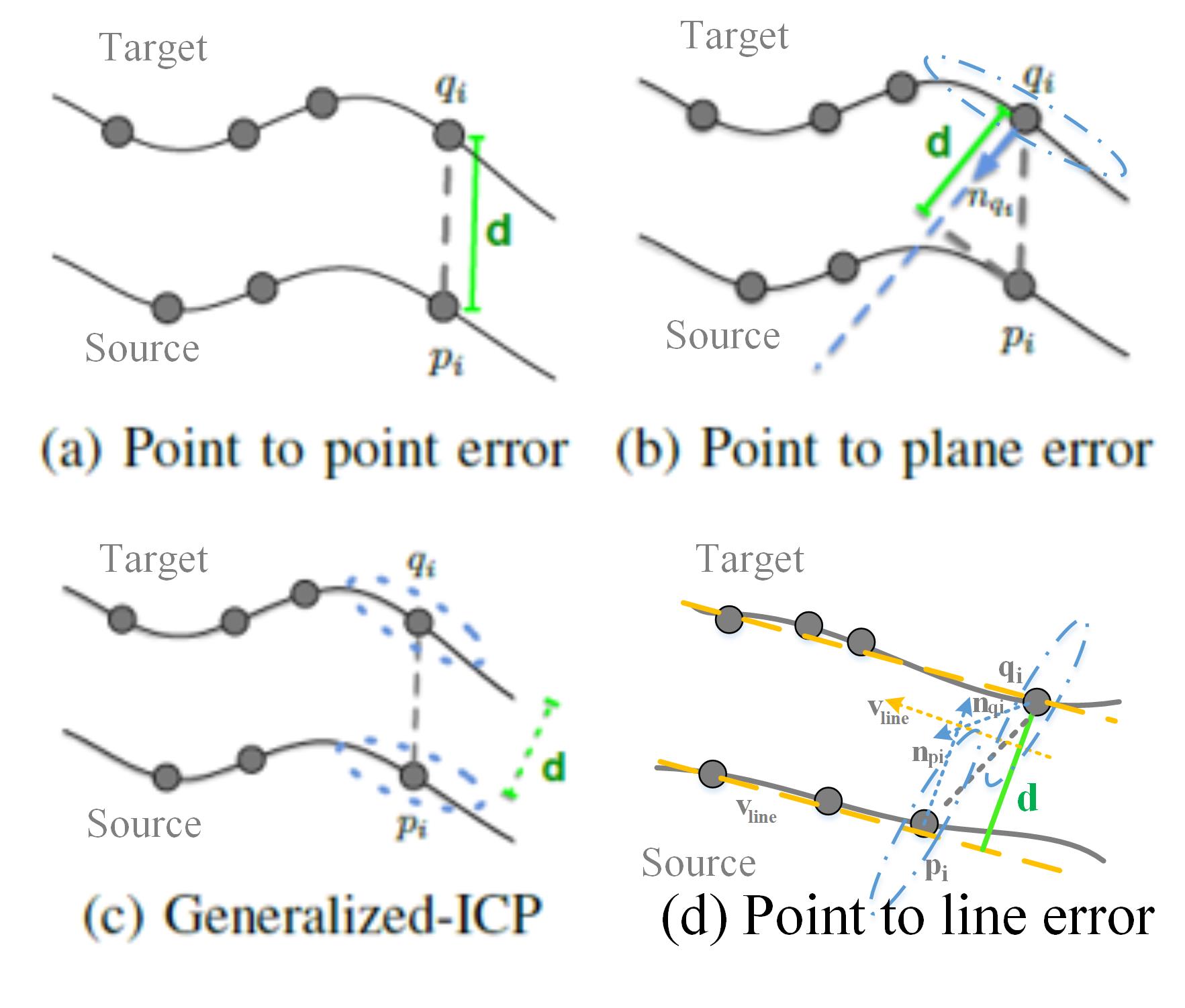}
\end{center}
   \caption{ICP variants with different distance metrics \cite{pclreg}.}
\label{img11}
\end{figure}

There are also some variants of ICP focus on the distance metrics \cite{pclreg} of transformation estimation target function as shown in Fig.\ref{img11}. In comparison with the point-to-point distance, the point-to-plane \cite{pointtoplaneicp}, point-to-line \cite{pointtolineicp} distance metrics have better performance on scenarios with plenty of facades (plane) or pillars (lines). Their target functions for transformation estimation are listed in Eq.\ref{equ15} and Eq.\ref{equ16} respectively, in which $n_{i}^{\left( p \right)}$ and $n_{i}^{\left( q \right)}$ are the normal vector of $p_i$ and $q_i$'s neighborhood. The idea of multiple distance metrics is applied in the state-of-art Lidar odometry solution LOAM \cite{loam}, which uses non-linear optimization to solve the point-to-plane and point-to-plane ICP. Furthermore, \cite{gicp} proposed the Generalized ICP, which adopts the neighborhood covariance matrix to combine different distance metrics together. For these methods, neighborhood Principle Component Analysis (PCA) needs to be done to get normal vector as well as neighborhood covariance.

\begin{equation}
\left\{ R^*,t^* \right\} =\underset{\left\{ R,t \right\}}{\text{arg}\min}\left( \sum_i{\lVert \left( Rp_i+t-q_i \right) \cdot n_{i}^{\left( q \right)} \rVert ^2} \right) 
    \label{equ15}
\end{equation}

\begin{equation}
\left\{ R^*,t^* \right\} =\underset{\left\{ R,t \right\}}{\text{arg}\min}\left( \sum_i{\lVert \left( Rp_i+t-q_i \right) \times \widehat{\left( n_{i}^{\left( q \right)}\times n_{i}^{\left( p \right)} \right) } \rVert ^2} \right) 
    \label{equ16}
\end{equation}

Since ICP tends to converged to wrong local optima with bad transformation initial guess\cite{icpreview}  as shown in Fig.\ref{img8}(b-c), 
other variants of ICP focus on broadening the basin of convergence and avoiding the local optimum. ICP with Invariant Features (ICPIF) \cite{icpif} combined invariant features with geometric distance in the closest distance calculation. ICPIF is more likely to converged to global optimum than ICP under ideal, noise free conditions. \cite{goicp} proposed the Global Optimal ICP (Go-ICP) to integrate ICP with a branch-and-bound (BnB) scheme so that a coarse registration is not needed. However, Go-ICP is much more time consuming than ICP and sensitive to outliers.

\subsection{Analysis and evaluation}

In conclusion, the advantages and limitations of ICP are listed as follows. On one hand, ICP is extremely dependent on a good initialization, without which the algorithm is likely to be trapped in a local optimum so that it's a kind of local registration method. On the other hand, ICP can achieve high registration accuracy when the rotation deviation to the ground truth is small so that ICP is often known as a preferred fine registration method. The general strategy to do registration for TLS point cloud is to apply coarse registration method at first and then use ICP to refine the coarse result. Besides, since ICP is somewhat efficient and following a simple and versatile processing structure, it is the most popular algorithm for SLAM related applications nowadays \cite{icpreview}.

\section{Randomized hypothesize-and-verify based algorithms}\label{sec5}

\subsection{RANSAC without global correspondences determination}
The most representative randomized hypothesize-and-verify based algorithms is RANSAC \cite{RANSAC}. Without doing correspondence determination (feature matching), we can also apply RANSAC without correspondence to find the largest common pointset for determining the correct registration. Randomly select three different points from the source point cloud and three from the target point cloud to form a  group of correspondence bases, estimate the candidate transformation that register the base pairs, and then count the number of point from transformed source point cloud that within a inlier distance threshold from the nearest points in the target point cloud. The transformation estimated by the base pairs with the most points within distance threshold would be accepted at last. 

The problem is the efficiency due to the equation of minimum iteration number for a  trustworthy sample set as shown in Eq.\ref{equ18}, in which $r$ is the point inlier ratio, $N$ is the sample number (6 here) , $0.99$ is the confidence and $M$ is the trial number.

\begin{equation}
p=1-\left( 1-r^N \right) ^M>0.99
    \label{equ18}
\end{equation}

The better solution is to pick base points randomly from source point cloud, and efficiently look for geometric congruent corresponding points in target point cloud. Though the time complexity for congruent group searching is $O(n^3)$, the sample number $N$ would decrease to 3 and the inlier ratio $r$ would increase a lot, thus leading to less total trial number.

\subsection{4 Point Congruent Sets (4PCS) algorithm}

\begin{figure}[!h]
\begin{center}
\includegraphics[width=0.9\linewidth]{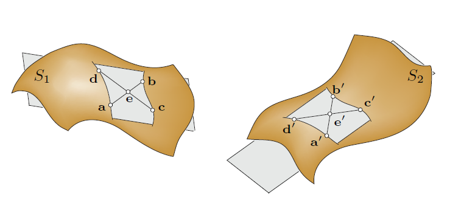}
\end{center}
   \caption{4PCS algorithm: an example of 4-points congruent set pair \cite{4PCS}.}
\label{fig:12}
\end{figure}

\begin{figure}[!h]
\begin{center}
\includegraphics[width=1.0\linewidth]{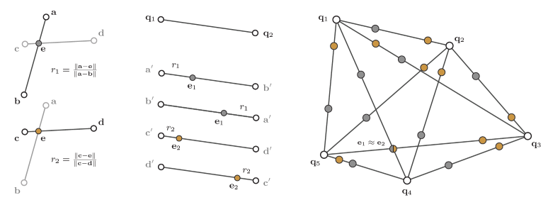}
\end{center}
   \caption{4PCS algorithm: an example of congruent set match using the invariant intersection distance ratio \cite{4PCS}.}
\label{fig:13}
\end{figure}

The 4-Points Congruent Sets (4PCS) \cite{4PCS} adopts the searching algorithm of affine transformation invariant coplanar 4 point group in different point sets with $O(n^2)$ time complexity \cite{fastaffine} to improve the congruent group searching efficiency since rigid transformation is the subset of affine transformation. 4PCS determines the corresponding four-point base sets by taking  advantage of the invariant intersection distance ratios of these four points, as shown in Fig.\ref{fig:12} and Fig.\ref{fig:13}.

Since 4PCS has good performance on challenging global registration cases but is still time-consuming due to the huge total number of points, some more efficient variants of 4PCS are proposed recently. Super4PCS \cite{Super4PCS} decrease the time complexity of congruent set searching from $O(n^2)$ to $ O(n)$ by using smart indexing of points. K-4PCS \cite{k4pcs} is based on significant keypoints instead of the raw point cloud so that decrease the processing point number.

Randomly hypothesis and verify strategy can also be used for plane correspondence. \cite{planarreg} use RANSAC for plane correspondence based registration. \cite{v4pcs} proposed V4PCS, which is a plane version of 4PCS. In V4PCS, plane primitive is fitted from each voxel. Then the invariant intersection angle between 4 plane set's normal vectors is adopted to speed up the congruent plane searching procedure.  

\subsection{Analysis and evaluation}
These random sample based algorithms also do not need the transformation initial guess so they can do global registration. Since  candidates are checked under a certain confidence, these methods are somehow robust to noise, outliers and similar (repetitive) structures but the total trial number would also increase under such circumstance because the likelihood of picking outlier free subsets degrades rapidly. Besides, these algorithm are mostly coarse registration solutions since the
final transform estimated from the pair base of only minimum required number of correspondence is not accurate enough.

\section{Summary and Outlook}\label{sec6}

Except for the aforementioned methods, there are also some probability based methods that don't follow the correspondence determination and transform estimation workflow. These probability based algorithms often fit some kind of probability distribution in target point cloud and then maximize the product of probability of the transformed points in source point cloud under certain distribution. Some examples are the 3D Normal Distribution Transformation (NDT) algorithm \cite{NDT}, the Coherent Point Drift (CPD) algorithm \cite{cpd} and the Gaussian Mixed Model Registration (GMMReg) algorithm \cite{GMMReg}. NDT has been widely used in LiDAR assisted localization. Since there's no exact one-to-one correspondence between two point clouds due to noise of measurement and sampling, applying the probability based strategy can better deal with such problem compared with correspondence based methods. However, other probability based methods are still not robust enough to handle registration on large scale real world scenarios. 

\begin{figure}[!h]
\begin{center}
\includegraphics[width=0.65\linewidth]{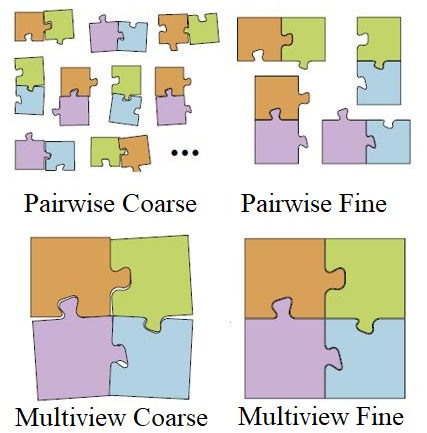}
\end{center}
   \caption{Common point cloud registration workflow: from pairwise to multi-view, from coarse to fine \cite{globallyreg}.}
\label{fig:14}
\end{figure}

Generally speaking, for practical registration cases with a lot of scans, we often follow the `coarse to fine' and `pairwise to multi-view' processing idea, as shown in Fig.\ref{fig:14}. Coarse registration algorithm like feature matching and 4PCS are used at first and then fine registration algorithm such as ICP and NDT are adopted to refine the former result. Then we usually use some global adjustment strategy like pose graph optimization \cite{globallyreg} \cite{comparereg} to jointly register multiple scans together and minimize the misclosure. 

Although there are so many different target-less registration methods that are suitable for different kind of datasets, they still face some common challenges, namely the huge number of points, low overlapping rate, the exist of clutters, occlusion, noise as well as the repetitive structures. Besides, the trade-off of accuracy and efficiency is still a big issue in practice. These challenges would be the main focus of future improvement of registration methods. 

There are some other open questions to solve in the near future. For example, the cross-platform registration (such as ALS and TLS point clouds) is still a challenging problem due to the huge difference in perspective ,range and  point density. Besides, the little-overlapping registration or shape matching problem which can be very useful in digital cultural relic restoration still waits for good solutions.

{\small
\bibliographystyle{ieee}
\bibliography{main.bib}
}

\end{document}